\documentclass{article}

\PassOptionsToPackage{numbers,compress}{natbib}
\usepackage[final]{neurips_2025}

\usepackage{neurips_2025}

\makeatletter
\renewcommand{\@notice}{} 
\makeatother

\usepackage{siunitx}
\usepackage{verbatim}
\usepackage{placeins}

\usepackage[utf8]{inputenc} 
\usepackage[T1]{fontenc}    
\usepackage{url}            
\usepackage{booktabs}       
\usepackage{textcomp}       
\usepackage{amsfonts}       
\usepackage{adjustbox}      
\usepackage{algcompatible}
\usepackage[noend]{algpseudocode}  
\usepackage{algcompatible}    
\usepackage{amsmath}          
\usepackage{algorithm}
\usepackage{amssymb}
\usepackage{enumitem}       
\usepackage{nicefrac}       
\usepackage{microtype}      
\usepackage{xcolor}         
\usepackage{graphicx}
\usepackage{siunitx}
\usepackage{subcaption}
\usepackage{etoolbox}       

\usepackage{hyperref}

\hypersetup{
    colorlinks=false,         
    pdfborder={0 0 1},        
    citebordercolor={0.4 1 0.4},  
    linkbordercolor={1 0 0}    
}

\usepackage[nameinlink,noabbrev]{cleveref}

\crefname{section}{Section}{Sections}
\Crefname{section}{Section}{Sections}
\crefname{subsection}{Section}{Sections}
\Crefname{subsection}{Section}{Sections}
\crefname{figure}{Figure}{Figures}
\Crefname{figure}{Figure}{Figures}
\crefname{table}{Table}{Tables}
\Crefname{table}{Table}{Tables}
\crefname{algorithm}{Algorithm}{Algorithms}
\Crefname{algorithm}{Algorithm}{Algorithms}
\crefname{appendix}{Appendix}{Appendices}
\Crefname{appendix}{Appendix}{Appendices}

\newcommand{\AppendixCrefSetup}{%
  \crefalias{section}{appendix}%
  \crefalias{subsection}{appendix}%
  \crefalias{subsubsection}{appendix}%
}
\pretocmd{\appendix}{\AppendixCrefSetup}{}{}

\renewcommand{\autoref}[1]{\cref{#1}}

\title{A Neuroscience-Inspired Dual-Process Model of Compositional Generalization}

\author{
  Alex Noviello\textsuperscript{1}\textsuperscript{\dag} \quad
  Claas Beger\textsuperscript{1}\textsuperscript{\dag} \quad
  Jacob Groner\textsuperscript{1} \quad
  Kevin Ellis\textsuperscript{1}\textsuperscript{\ddag} \quad
  Weinan Sun\textsuperscript{2}\textsuperscript{\ddag} \\
  \textsuperscript{1}Department of Computer Science, Cornell University \\
  \textsuperscript{2}Department of Neurobiology and Behavior, Cornell University \\
  \texttt{\{abn52,cbb89,jrg349,kellis,ws467\}@cornell.edu}
}

\begin{document}

\maketitle

\begingroup
\renewcommand\thefootnote{\dag}\footnotetext{Equal contribution.}
\renewcommand\thefootnote{\ddag}\footnotetext{Equal advising.}
\endgroup


\begin{abstract}
Deep learning models struggle with systematic compositional generalization, a hallmark of human cognition. We propose \textsc{Mirage}, a neuro-inspired dual-process model that offers a processing account for this ability. It combines a fast, intuitive ``System~1'' (a meta-trained Transformer) with a deliberate, rule-based ``System~2'' (a Schema Engine), mirroring the brain's neocortical and hippocampal--prefrontal circuits. Trained to perform general, single-step decomposition on a stream of random grammars, \textsc{Mirage} achieves $>$99\% accuracy on all splits of the SCAN benchmark in a task-agnostic setting. Ablations confirm that the model's systematic behavior emerges from the architectural interplay of its two systems, particularly its use of explicit, prioritized schemas and iterative refinement. In line with recent progress on \emph{recursive/recurrent} Transformer approaches, \textsc{Mirage} preserves an iterative neural update while \emph{externalizing} declarative control into an interpretable schema module. Our work provides a concrete computational model for interpreting how compositional reasoning can arise from a modular cognitive architecture.
\end{abstract}

\section{Introduction}

Humans effortlessly generalize by composing known concepts in new ways, a capacity termed systematic compositionality~\cite{fodor1975language,lake2017building}. Yet, even large-scale neural networks often fail at such generalization, tending to learn shallow heuristics instead of abstract rules~\cite{fodor1988connectionism,lake2018generalization,xu2024large}. This paper provides a processing account for compositional reasoning, inspired by the dual-process architecture of the human brain.

Cognitive neuroscience suggests that intelligent behavior arises from the interplay of complementary learning systems (CLS)~\cite{mcclelland1995there,kumaran2016learning}. This theory posits that a fast, intuitive ``System 1,'' supported by the neocortex, handles pattern recognition, while a slower, deliberate ``System 2,'' involving the hippocampus (HPC) and prefrontal cortex (PFC), supports structured, episodic, and goal-directed reasoning~\cite{moscovitch2016episodic,bein2025schemas}. The HPC rapidly encodes specific experiences (episodes), which the PFC abstracts over time into reusable schemas, generalizable knowledge structures, that enable flexible planning and inference~\cite{tse2007schemas,preston2013interplay}. This division of labor allows for both rapid adaptation and systematic application of knowledge.

We instantiate this theory in \textsc{Mirage} (Meta-Inference with Rules and Abstractions from Generalized Experience), a dual-process model with two interacting components:
\begin{enumerate}[label=(\arabic*), topsep=2pt, itemsep=1pt, parsep=1pt]
    \item \textbf{System~1 (Neural Decomposer)}: A compact Transformer trained via meta-learning to perform a single step of compositional decomposition. This parallels the fast, pattern-matching function of the neocortex.
    \item \textbf{System~2 (Schema Engine)}: An explicit symbolic module that extracts, stores, and applies prioritized schemas, managing variable bindings in an episodic memory. This mirrors the deliberate, schema-based reasoning of the HPC--PFC loop.
\end{enumerate}
At inference, these systems engage in an iterative loop: System 2 extracts a compositional grammar from examples and uses it to guide the repeated application of System 1. Each pass of the System 1 Transformer refines the problem representation by one level, allowing the model to recursively decompose complex structures. This architecture provides an interpretable mechanism for generalization. On the SCAN benchmark~\cite{lake2018generalization}, \textsc{Mirage} achieves state-of-the-art performance across all splits, despite its Transformer component never being trained on SCAN-specific data. Our work demonstrates that a modular, neuro-inspired architecture can provide a powerful and interpretable processing account for systematic generalization.

\begin{figure}[t]
    \centering
    \includegraphics[width=\linewidth]{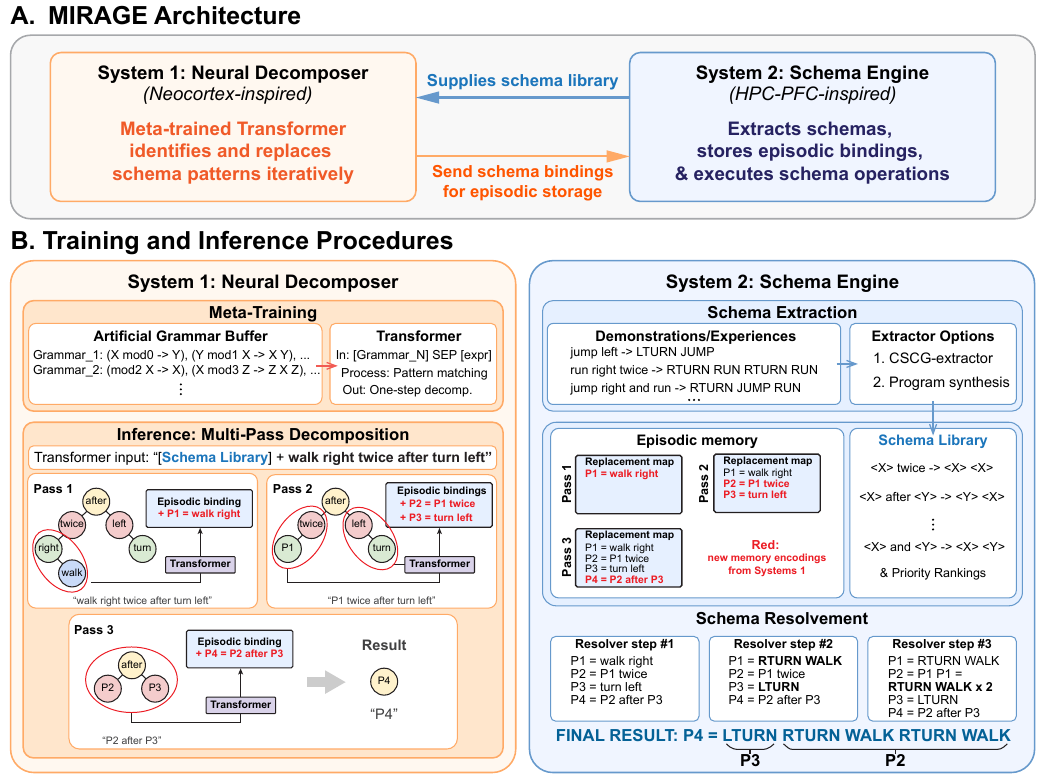}
    \caption{\textsc{Mirage} architecture and inference loop. (A) System 1 (Neural Decomposer) is a meta-trained Transformer for pattern matching. System 2 (Schema Engine) extracts and manages a library of prioritized schemas, modeling HPC--PFC function. (B) During inference, the systems iterate. For a command like ``walk right twice after turn left,'' System 2 provides the relevant grammar. System 1 applies one decomposition step per pass, with outputs stored in an episodic memory. This process repeats, reducing the composition tree layer by layer until a primitive action sequence is produced.}
    \label{fig:Architecture}
\end{figure}

\section{A Dual-Process Framework}
\label{sec:method}

\textsc{Mirage} solves compositional tasks through an iterative refinement process managed by its two systems. The core idea is to offload declarative knowledge (the "what") to an explicit schema library (System 2) and procedural knowledge (the "how") to a general-purpose neural reasoner (System 1).

\paragraph{System 2: Schema Engine.} A compositional task is defined by a grammar $\mathcal{G} = (\Sigma,\pi, \psi)$, where $\Sigma$ is a set of schemas (rewrite rules), $\pi$ is a priority ordering, and $\psi$ is an evaluator for primitive actions. System 2's role is to induce this grammar from a small set of examples. This extraction is modular; we show it can be done with either a symbolic rule-miner or a neuro-inspired model based on Clone-Structured Causal Graphs (CSCGs)~\cite{George2021-bo,sun2025learning}. For details on these methods, see \autoref{subsec:enumerative_rule_miner} and \autoref{sec:CSG_Extractor_apx}.

\paragraph{System 1: Meta-Learned Transformer Decomposer.} System 1 is a decoder-only Transformer $\mathcal{T}_{\theta}$ trained not to solve any single task, but to perform a generic, single step of compositional decomposition. We meta-learn $\mathcal{T}_{\theta}$ on an endless stream of randomly generated grammars. Each training sample consists of a grammar definition, an input sequence, and the target output after one decomposition step at the highest-priority location. This forces the model to learn the abstract *process* of rule application rather than memorizing solutions for a specific task. See \autoref{A1transformervocab} and \autoref{A2trainingalogs} for a full description of the training procedure and model vocabulary.

\paragraph{Inference as a Processing Account.} At inference time, System 2 first extracts the task grammar $\mathcal{G}$. Then, for a given input command, the model enters a loop (\autoref{fig:Architecture}B) that serves as a processing account of step-by-step deliberation. In each step, the current sequence and the grammar are fed to the Transformer $\mathcal{T}_{\theta}$. For example, given `walk right twice after turn left' and a grammar where `twice' has higher priority than `after', the Transformer identifies `walk right twice' and rewrites it to a placeholder token, say \texttt{P1}. The sequence becomes `\texttt{P1} after turn left'. In the next iteration, it decomposes this simpler structure. This process repeats until only a sequence of primitive actions remains. This iterative refinement allows a small model with a fixed context window to solve problems of arbitrary compositional depth. The full inference algorithm is detailed in \autoref{alg:inference}.

\section{Experiments and Behavioral Analysis}
\label{sec:results}
We evaluate \textsc{Mirage} on the canonical \textbf{SCAN} benchmark~\cite{lake2018generalization}, which tests systematic generalization to novel compositions and longer sequences. The System 1 Transformer (1.19M parameters) is trained task-agnostically on random grammars; for evaluation, the System 2 Schema Engine extracts the SCAN grammar from its training split, which is then provided to the frozen Transformer.

\subsection{Compositional Generalization Performance}
As shown in \autoref{tab:scan_performance}, \textbf{\textsc{Mirage} achieves near-perfect accuracy on all SCAN splits}, including those designed to foil standard sequence-to-sequence models. A vanilla Transformer trained directly on SCAN's simple split masters it but completely fails on the generalization splits, indicating it learns superficial correlations rather than abstract rules. Providing the schema library as a simple prompt (`Transformer+SC\_Library') does not help, confirming that a specialized architecture and training objective are required to properly utilize schematic knowledge.

\begin{table}[h]
  \caption{SCAN accuracy (mean\,\,$\pm$\,SEM over 4 runs). Baselines are re-trained on each split's training set. \textsc{Mirage} uses a single, task-agnostically trained Transformer for all splits.}
  \label{tab:scan_performance}
  \centering
  \small
  \setlength{\tabcolsep}{4pt}%
  \begin{tabular*}{\linewidth}{@{\extracolsep{\fill}} l c c c c c}
    \toprule
    & & \multicolumn{4}{c}{\textbf{SCAN splits}}\\
    \cmidrule(lr){3-6}
    \textbf{Model} & \textbf{Full Task} & \textbf{Simple} & \textbf{Length} & \textbf{Add prim. `jump'} & \textbf{Template} \\
    \midrule
    Transformer &
      N/A &
      $99.85${\scriptsize$\,\pm\,0.00$} &
      $13.58${\scriptsize$\,\pm\,0.01$} &
      $0.40${\scriptsize$\,\pm\,0.13$} &
      $3.09${\scriptsize$\,\pm\,3.01$} \\
    Transformer\tiny{+SC\_Library} &
      N/A &
      $99.91${\scriptsize$\,\pm\,0.11$} &
      $15.86${\scriptsize$\,\pm\,1.36$} &
      $0.03${\scriptsize$\,\pm\,0.04$} &
      $0.00${\scriptsize$\,\pm\,0.00$} \\
    \textsc{Mirage} (ours) &
      $99.59${\scriptsize$\,\pm\,0.24$} &
      $99.50${\scriptsize$\,\pm\,0.30$} &
      $99.35${\scriptsize$\,\pm\,0.41$} &
      $99.65${\scriptsize$\,\pm\,0.20$} &
      $99.55${\scriptsize$\,\pm\,0.23$} \\
    \bottomrule
  \end{tabular*}
\end{table}

\subsection{Processing Account: Ablation Studies}
To understand which components drive this behavior, we conducted several ablations. These studies confirm that \textsc{Mirage}'s performance is an emergent property of its cognitive architecture, not just one component.\newpage
\begin{enumerate}[itemsep=2pt,leftmargin=1.3em]
    \item \textbf{Schema priorities are crucial for cognitive control.} Removing the explicit priority ordering from the schema library causes accuracy to plummet from 99.6\% to 71.9\%. The model becomes unable to resolve ambiguous commands (e.g., `turn left twice and walk'), analogous to a failure of executive function in selecting the correct order of operations.
    \item \textbf{Iterative refinement enables deliberation.} Replacing the iterative loop with a single-pass, end-to-end mapping fails to generalize. The step-by-step process is essential, allowing the model to focus on one sub-problem at a time, a form of deliberate, sequential reasoning that monolithic models lack.
    \item \textbf{Accurate schemas are necessary.} Introducing minor errors into the extracted grammar (e.g., misidentifying a schema's arguments) causes performance to collapse to nearly 0\%. This highlights the model's reliance on accurate declarative knowledge. Its reasoning algorithm is sound, but it cannot function with a corrupted world model, a behavior consistent with human reasoning \cite{DelusionBeliefs}.
\end{enumerate}

\section{Related Work}
Research on compositionality spans auxiliary objectives, neuro-symbolic models, meta-learning, and analyses of Transformer mechanisms. Auxiliary supervision can encourage structural understanding in Transformers \cite{jiang2021inducingtransformerscompositionalgeneralization}, while memory-augmented models like LANE \cite{liu2020compositionalgeneralizationlearninganalytical} explore variable-slot reasoning. Neuro-symbolic systems such as the Compositional Program Generator \cite{klinger2024compositionalprogramgenerationfewshot} and the Neural-Symbolic Recursive Machine \cite{li2024neuralsymbolicrecursivemachinesystematic} highlight the benefits of grammar-based modularity, paralleling \textsc{Mirage}’s schema-based design. 

Meta-learning approaches have also demonstrated strong potential for systematic generalization. In particular, \citet{lake2023human} show that a meta-learned neural network can achieve human-like compositional generalization, supporting the view that inductive biases learned from experience can yield flexible reasoning skills.

Transformer-focused studies reveal both strengths and limits: while large models can sometimes achieve compositional generalization \cite{russin2024fregechatgptcompositionalitylanguage}, their reasoning often emerges only through grokking \cite{wang2024grokked} or limited mechanistic patterns \cite{brinkmann2024mechanisticanalysistransformertrained}, underscoring the need for specialized architectures.

Finally, neuroscience perspectives propose that hippocampal replay supports compositional computation \cite{kurthnelson2022replaycompositionalcomputation}, and that HPC–PFC circuits represent an evolutionary solution to flexible reasoning \cite{murray2017evolution}. These ideas motivate \textsc{Mirage}’s dual-process design, bridging cognitive neuroscience and AI.

\section{Discussion and Conclusion}
\label{sec:conclusion}
\textsc{Mirage} offers a hybrid approach to compositional generalization, using a neural network for a general procedural task (decomposition) while relying on an explicit, interpretable symbolic layer for declarative knowledge and control. Unlike monolithic models that must implicitly learn reasoning algorithms through “grokking,” \textsc{Mirage} explicitly encodes a process of deliberate, iterative refinement. This dual-process framework offers a concrete processing account for systematic generalization. By instantiating cognitive theories of complementary learning systems, our model achieves state-of-the-art zero-shot performance on SCAN. We release the full implementation of \textsc{Mirage} on \href{https://github.com/ClaasBeger/MIRAGE}{\textsc{GitHub}}. Our main contributions as an interpretive work are:
\begin{itemize}[leftmargin=*,topsep=2pt,itemsep=1pt]
    \item \textbf{Processing Account:} We show that systematicity can emerge from an architecture that separates procedural skill from declarative knowledge and applies them iteratively.
    \item \textbf{Behavioral Account:} The model's success where standard Transformers fail demonstrates the behavioral advantage of a neuro-inspired, modular design.
    \item \textbf{Developmental Account:} Meta-learning provides a hypothesis for how agents can acquire robust, generalizable cognitive algorithms from varied experience.
    \item \textbf{Interpretability Account:} Because \textsc{Mirage} explicitly marks and resolves schemas step by step, it produces a transparent trace of intermediate problem states. This contrasts with black-box models that leap directly to an answer, making our framework inherently more interpretable and diagnostically useful.
\end{itemize}

\newpage

\paragraph{Connection to Cognitive Learning Systems}
Beyond enabling stepwise compositional inference, schema formation as utilized by \textsc{Mirage} provides a mechanism for rapid assimilation of new rules into structured knowledge. In neuroscience, once an associative schema is established, new information consistent with that schema is integrated and becomes neocortically supported unusually quickly \citep{tse2007schemas,preston2013interplay,kumaran2016cls}. Recent work further links medial prefrontal schema representations with reinforcement-learning style updating and deployment for flexible behavior \citep{bein2025schemas}. \textsc{Mirage}'s prioritized schema selection echoes this PFC-mediated control over hippocampal content during retrieval and planning \citep{preston2013interplay}.

In addition, when viewed through a continual-learning lens, schemas could act as structured inductive priors: previously consolidated structure may be used to constrain and accelerate learning on new tasks. Bayesian treatments formalize this as using posteriors from past tasks to define priors for future ones, while also highlighting practical limits when approximations are poor \citep{farquhar2019unifying,kessler2023sequential}. These observations motivate future work in the direction of probabilistic extension: inferring over candidate schema targets (and, when needed, over the library itself) to maintain sample-efficient generalization when direct schema identification is challenging.

\paragraph{Relation to recurrent/recursive Transformers}
A growing line of work explores \emph{iterative} neural reasoning with weight-tying over steps, truncated credit assignment, and, in some cases, adaptive halting, yielding recurrent or recursive Transformer loops \citep{dehghani2018universal,bai2019deep,wang2025hierarchical,jolicoeurmartineau2025trm}. \textsc{Mirage} lies squarely within this trend by retaining an iterative per-step neural update, yet it differs in where the \emph{control signal} lives: rather than learning control implicitly, we \emph{externalize} it as explicit schemas with priorities. This keeps the empirical benefits of looped refinement while producing stepwise, editable traces and an algorithmic-level account of the reasoning process, which are central to Cognitive Interpretability. In this sense, \textsc{Mirage} complements recurrent and recursive Transformer practice with a cognitively motivated control layer, clarifying how and why each step is chosen.

\paragraph{Limitations and Future Work}
While successful on SCAN, the current framework relies on reliable extraction of a consistent grammar. Future work should explore robustness to noisy or incomplete schemas. Such issues could potentially be addressed by extending the extractor with probabilistic rule learning and/or the Neural Decomposer with active retrieval mechanisms. We also plan to apply this architecture to more complex reasoning tasks and use its inspectable states to test specific neuroscientific hypotheses about HPC--PFC interactions during planning. By building explicit, interpretable cognitive architectures, we can move beyond black-box evaluations and toward a mechanistic understanding of reasoning in neural systems.

\clearpage
{\small \bibliography{references}}

\clearpage

\appendix

\section{Methodology details}
\label{sec:methodology_details}

\subsection{Formal definition of Transformer Inference}
\begin{algorithm}[htb]
\caption{Zero-Shot Inference on Single Input Sequence Example}
\label{alg:inference}
\begin{algorithmic}[1]
\Require Trained transformer $\mathcal{T}_\theta$, grammar $\mathcal{G}$, input sequence $x^{(k)}$ with composition depth $k$
\Ensure Final flat output sequence $\mathcal{S}_{\mathrm{final}}$ and replacement map $\lambda$
\State $\lambda \gets \{\}$
\Comment{empty map from schema-instances to primitives}
\State $x^{(k)} \gets$ input sequence
\For{$d = k, k-1, \dots, 1$}
  \State $y \gets \mathcal{T}_\theta\bigl(x^{(d)}\bigr)$
  \Comment{predict next-level decomposition}
  \While{there is a schema token $s$ in $y$}
    \State let $(\sigma_s,\,a_1,\dots,a_m)$ be the schema name and its $m$ argument tokens in $y$
    \State $p \gets \texttt{ApplySchema}(\sigma_s,\,a_1,\dots,a_m)$
    \Comment{collapse into one primitive}
    \State $\lambda\bigl[\sigma_s(a_1,\dots,a_m)\bigr] \gets p$
    \State replace the subsequence $\langle \sigma_s,a_1,\dots,a_m\rangle$ in $y$ with $p$
  \EndWhile
  \State $x^{(d-1)} \gets y$
  \Comment{one less level of composition}
\EndFor
\State \Return $\mathcal{S}_{\mathrm{final}} \gets x^{(0)},\;\lambda$
\end{algorithmic}
\end{algorithm}

\subsection{Formal definition of CSCG-based System 2 Algorithm}
\begin{algorithm}[htb]
\caption{Extraction of compositional schemas}
\label{alg:schema}
\begin{algorithmic}[1]
\Require\ Demonstrations $\mathcal{S}$, minimum support $k$
\State Train CHMM on $\mathcal{S}$; decode each $s\in\mathcal{S}$ to obtain episodes $\mathcal{E}$
\State $\mathcal{C}\gets\varnothing$ \Comment{candidate set}
\ForAll{$(e_i,e_j)\in\binom{\mathcal{E}}{2}$}
  \State $\tau\gets\textsc{Align}(e_i,e_j)$ \Comment{injective LCS with variables}
  \If{$\tau\not=\bot$}
     \State $\text{supp}\gets\{e\in\mathcal{E}\mid\textsc{Validate}(\tau,e)\}$
     \If{$|\text{supp}|\ge k$} $\mathcal{C}\gets\mathcal{C}\cup\{\langle\tau,\text{supp}\rangle\}$ \EndIf
  \EndIf
\EndFor
\State $\mathcal{S}_{\text{raw}}\gets\textsc{Deduplicate}(\mathcal{C})$
\State $\mathcal{S}_{\text{final}}\gets\textsc{PruneByComposition}(\mathcal{S}_{\text{raw}})$
\State \Return $\mathcal{S}_{\text{final}}$
\end{algorithmic}
\end{algorithm}

\subsection{Transformer Model Vocabulary Specifications for Meta-Learning}
\label{A1transformervocab}
The vocabulary of our transformer $\mathcal{T}_{\theta}$ is defined with the following components. To use the model with any arbitrary grammar with its own specific vocabulary, like SCAN, tokens must be anonymized to match this format. For example, in SCAN, `primitives' are tokens like `walk', `jump', `look', etc., while modifiers are tokens `and', `after', `opposite left', or `twice'. Anonymizing a grammar in this way is a simple, deterministic process.
\begin{enumerate}
    \item \textbf{Primitives}: \texttt{PRIM\_0}, \dots, \texttt{PRIM\_P}. Primitives are the basic atomic elements in a vocabulary. Integer parameter $P$ sets the number of allowed unique primitives.
    \item \textbf{Modifiers}: \texttt{MOD\_0}, \dots, \texttt{MOD\_M}. Schemas bind to and define the action of modifier tokens as functions. A given grammar $\mathcal{G}$ defines the action of a modifier in a sequence generated under it. Integer parameter $M$ sets the number of allowed unique modifiers.
    \item \textbf{Argument Tokens}: \texttt{ARG\_0}, \dots, \texttt{ARG\_{2*A}}. Argument tokens are used in the in-context token sequences used to represent grammars. Specifically, argument tokens define the placeholders that a schema could bind to. For example, a schema may be represented presented via a token sequence as \texttt{ARG\_0, \dots, ARG\_A MOD\_0 ARG\_A+1, \dots, ARG\_2A}, where the argument tokens are placeholders for real arguments to the schema and the modifier token represents the action the schema is binding to. Integer parameter $A$ sets the maximum number of arguments a schema can have that occur before and after its principle modifier.
    \item \textbf{Schema Name Tokens}: \texttt{SC\_0, \dots, SC\_S}. These tokens define the names of specific schemas in the grammar or $|\Sigma|$. These names occur in-context to serve as markers for bindings in output sequences after a model application. Integer parameter $S$ controls the number of schemas in a grammar, or $|\Sigma|$. We must have $S \geq m$ for a grammar G, as each modifier must occur in at least one schema to be well-defined.
    \item \textbf{Priority Tokens}: \texttt{PRIORITY\_0}, \dots, \texttt{PRIORITY\_S}. These tokens define the priority with which schemas are to be bound to. These tokens occur next to schema name tokens in-context to define their priority.
    \item \textbf{Administrative Tokens}: \texttt{EOS}, \texttt{SEP}, \texttt{SC\_DEF}, \texttt{SC\_PRI}, \texttt{SC\_SEP}, \texttt{LP\_SEP} , \texttt{PAD}. These tokens are for formatting grammars in-context with input sequences generated by them.
\end{enumerate}

\subsection{Training Algorithms}
\label{A2trainingalogs}
During the transformer training process, new random grammars are generated at regular intervals. These grammars are added to a `Grammar Buffer' maintaining the full set of previously generated grammars during the training process. At every step, the transformer samples grammars from this buffer and delegates the construction of random 2-deep composition input sequences and their corresponding output sequences. The concatenation of these components is then fed into the model. Multiple grammars are generally represented in each batch.

In the Methods section, we referenced a variety of small sub-algorithms that the transformer must delegate during the meta-learning training process. These include actually generating random grammars to add to the buffer at regular intervals (defined by a hyperparameter), generating input/output sequence pairs given a grammar, and actually sampling from the grammar buffer for each batch. Each of these algorithms are detailed here.
\begin{itemize}
  \item \textbf{Random Grammar Generation:} For some subset of modifier tokens $M_G \subseteq  M$, we generate a schema, $\sigma$. For each,
  \begin{itemize}
       \item Choose $A_{before}, A_{after} \leq A_{max}$ for number of arguments before and after the modifier token in the schema output format.
       \item Define $\psi(\sigma, (t_1, \dots, t_{A_{before} + A_{after}}))$.
       \item Define $\pi(\sigma)$.
     \end{itemize}
  \item \textbf{Input Sequence Generation:} To generate an input sequence from a grammar, choose a schema, $\sigma$. For each argument in $\sigma$, select an atomic primitive or another schema $\sigma_{sub}$ with $\pi(\sigma) < \pi(\sigma_{sub})$. For $\sigma_{sub}$'s, select all primitives as arguments. This generates 2-deep schema composition sequences.
  \item \textbf{Output Sequence Generation:} For the deepest layer of composition, simply replace the modifier token $M$ of applied schema $\sigma_M$ with a 'schema name token' like \texttt{SC\_M} to represent $\sigma_M$. We train our model to output this format so that we can easily detect where the model has performed decomposition. This facilitates the zero-shot inference procedure, given a new grammar specification, outlined below.
  \item \textbf{Grammar Buffer Sampling:} A Grammar Buffer consists of a set of $N$ previously used grammars. The buffer maintains the content of these grammars, as well as a corresponding $C_i$ representing the number of times a grammar $G_i$ has been used. When sampling from the buffer, we sample from the inverse $C_i$'s for all buffers. Let $C'_i$ be $\frac{1}{C_i + s}$, for smoothing factor $s, 0 < s \leq 1$. Then, the probability of selecting $G_i$ is $\frac{C'_i}{\sum_{j=1}^{N}C'_j}$.
\end{itemize}

\FloatBarrier
\clearpage
\section{Additional details about schema extractors}
\label{sec:schema_extractors}

\subsection{Priority Scoring Algorithm for CSCG-extractor}
\label{alg:precedence_apx}
\begin{algorithm}[htbp]
\caption{Learning operator precedence from one-step demonstrations}
\label{alg:precedence_simple}
\begin{algorithmic}[1]
\Require fixed schema set $\mathcal{S}$, demonstration episodes $\mathcal{E}$ of the form $\langle\text{cmd}\rangle\!<\text{SEP}>\!\langle\text{1-step}\rangle$
\State $C\gets\mathbf{0}$ \Comment{pair-wise win counts}
\ForAll{episode $e\in\mathcal{E}$}
    \State $(\mathbf{x},\mathbf{y})\leftarrow$ split $e$ at \texttt{<SEP>}
    \State $P \gets \{\,s\in\mathcal{S}\mid\text{pattern}(s)\subset\mathbf{x}\,\}$ \Comment{schemas present in the input command}
    \State $F \gets \{\,s\in P\mid\text{apply}(s,\mathbf{x})=\mathbf{y}\,\}$ \Comment{schemas that fired in the first step}
    \ForAll{$s_{w}\in F$}
        \ForAll{$s_{\ell}\in P\setminus F$}
            \State $C[s_{w},s_{\ell}] \gets C[s_{w},s_{\ell}] + 1$
        \EndFor
    \EndFor
\EndFor
\ForAll{$s_i\in\mathcal{S}$}
    \State $\displaystyle \text{score}(s_i)\gets\sum_{j\ne i} \bigl([\![C_{i,j}>0]\!]-[\![C_{j,i}>0]\!]\bigr)$ \Comment{Copeland score}
\EndFor
\State \Return $\{\text{score}(s_i)\}_{s_i\in\mathcal{S}}$
\end{algorithmic}
\end{algorithm}

\subsection{Enumerative Rule-Miner Details}
\label{subsec:enumerative_rule_miner}

This appendix expands on the \emph{enumerative rule miner} introduced in \autoref{sec:method}.

\subsubsection{Goal}
From a support set $\mathcal D=\{(\mathbf{x}^{(i)},\mathbf{y}^{(i)})\}_{i=1}^{N}$ of input–output strings, the miner induces a rewrite grammar $G=(\mathcal P,\mathcal M,\Sigma,\pi)$: primitives $\mathcal P$, modifiers $\mathcal M$, schemas $\Sigma=\{\sigma_1,\dots,\sigma_S\}$, and a precedence map $\pi$. Each schema is a variable-binding template whose left-hand side (LHS) may contain string literals, span variables $x_k$ (arbitrary substrings), or token variables $u_k$ (single words).

\subsubsection{Algorithm}
\begin{enumerate}[leftmargin=1.2em,label=(\arabic*)]
    \item \textbf{Template generation} Enumerate candidate LHS–RHS templates in order of description length via three language-agnostic edit primitives: (i) span $\!\to\!$ token replacements, (ii) span wrappers inserting a control token, and (iii) span splicing / re-ordering.
    \item \textbf{Repair test} Insert a candidate template $\tau$ into the current grammar and re-evaluate all demonstrations. $\tau$ is accepted \emph{iff} it rewrites every match consistently and increases corpus-level exact accuracy; accepted templates are appended to $\Sigma$.
    \item \textbf{Precedence induction} During replay, whenever two schemas match the same string and $\sigma_i$ fires before $\sigma_j$, record $\sigma_i\!\succ\!\sigma_j$. A topological sort of this graph yields $\pi$.
    \item \textbf{Termination} Iterate steps (1–3) until the candidate queue empties, accuracy plateaus, or a fixed budget is reached.
\end{enumerate}

Any alternative proposal mechanism (e.g., beam search, neural scorers, constraint solvers) can replace the enumerator in step (1) without changing the downstream interface: the Schema Engine consumes only $(\mathcal P,\mathcal M,\Sigma,\pi)$.

\section{CSCG application on SCAN}
\label{sec:cscg_application}
\begin{figure}[htbp]
    \centering
    \includegraphics[width=0.55\textwidth]{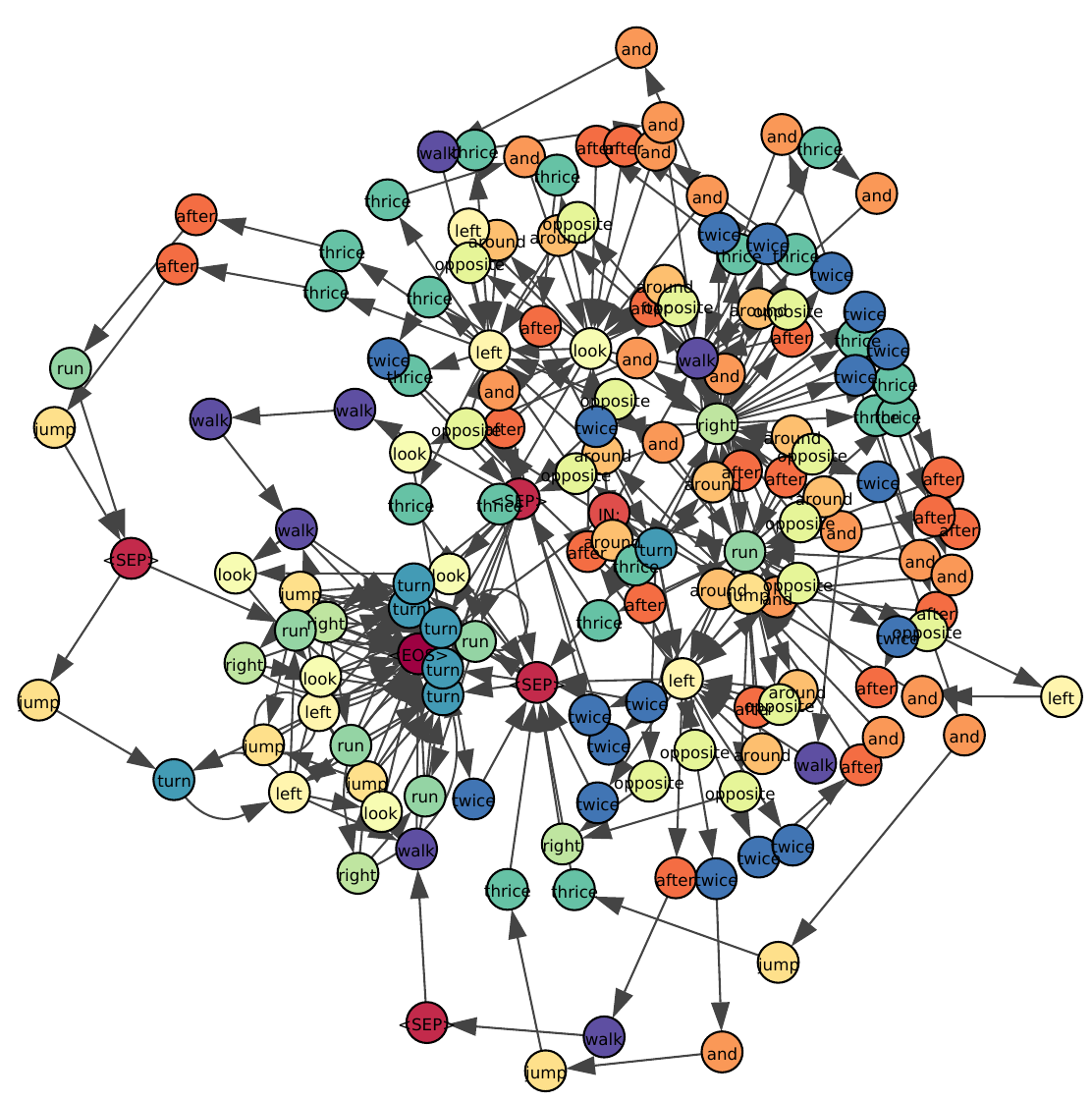}
    \caption{We apply a Clone-Structured Causal Graph with 100 clones directly on a concatenated subset of SCAN sequences and visualize the resulting model as a simple directed graph.}
    \label{fig:cscg_scan}
\end{figure}

In line with showcasing the inability of pure Transformers to solve different SCAN splits on its own in \autoref{sec:results}, we explored directly applying CSCG on concatenated SCAN sequences. However, due to innate traits of the model architecture, SCAN, or other compositional tasks, quickly exhaust the structural clone bottlenecks. Specifically, the model is unable to distinguish between more than n sequences due to the Separator token, which is contained between every input and output. At the point of entering this token, the model must commit to one specific clone, leading to the loss of prior information. We further experienced with alternative variants that take the state across all clones of a single emission into account, increasing the number of potential configurations that the state can exhibit, but these modifications still failed to meaningfully solve any SCAN split.

\section{Compute Resources}
\label{sec:compute}

Experiments were performed on readily available research hardware: single recent GPUs (e.g., A100 or H100) on a campus cluster and, occasionally, cloud services. Meta-training the Transformer required a few GPU-hours on one card, and each extra seed or ablation consumed a similar budget. Both schema extractors finish in under a minute on a CPU, and evaluating the full SCAN test set completes in under five minutes on a single GPU.

\section{CSCG Extractor Schemas}
\label{sec:CSG_Extractor_apx}

\begin{figure}[htbp]
    \centering
    \includegraphics[width=0.9\textwidth]{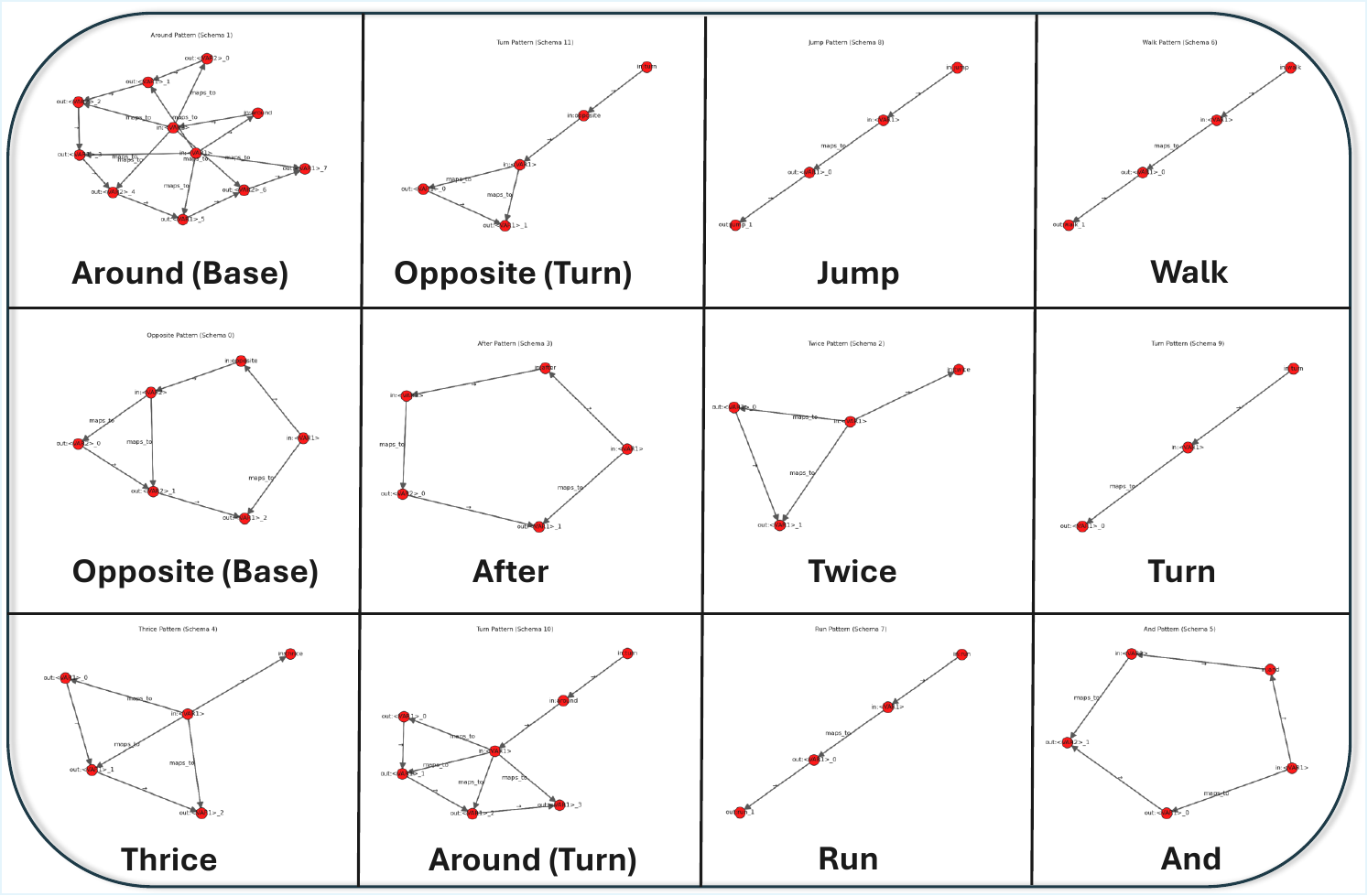}
    \caption{Overview of extracted schemas by the CSCG extractor, visualized as directed sequence graphs. Note that turn is a special case, which evokes different behavior when combined with around or opposite.}
    \label{fig:cscg_extract}
\end{figure}

The CSCG extractor produces both, a textual representation of the extracted schemas, as well as a graph-based visualization, showcasing the direct correspondence between input and output variables. This visualization approach reveals potential for schema comparison through graph-based representations, as atomic schema demonstrations share consistent structural components. Additionally, the extractor correctly identifies special cases in the "turn" and "around" schemas, recognizing that these produce no additional output when paired with the turn primitive, unlike other primitives such as jump or run.

\end{document}